\documentclass[letterpaper, 10 pt, conference]{ieeeconf}      % Comment this line out
\IEEEoverridecommandlockouts                             % This command is only
\usepackage[utf8]{inputenc}
\usepackage{graphicx}
\usepackage{epsfig}
\usepackage{amsmath}
\usepackage{amssymb} 
\usepackage{longtable}
\usepackage{rotating}
\usepackage{multirow}
\usepackage{array}
\usepackage{graphicx}
\usepackage{mathrsfs}
\usepackage{verbatim}
\usepackage{color}
\usepackage{float}
\usepackage{epstopdf}
\usepackage{url}
\usepackage{graphicx}
\usepackage{mathrsfs}
\usepackage{verbatim}
\usepackage{soul}
\usepackage{algorithm}
\usepackage{algorithmicx}
\usepackage{algorithm}
\usepackage[noend]{algpseudocode}
\usepackage[noadjust]{cite}
\usepackage{epstopdf}
\usepackage{url}
\usepackage{pgfplots,tikz-3dplot}
\usepackage{amsmath} % assumes amsmath package installed
\usepackage{amssymb}  % assumes amsmath package installed
\usepackage{bbm}
\usepackage{color}
\usepackage{graphics}
\usepackage{graphicx}
\usepackage{epsfig}
\usepackage{epstopdf}
\usepackage{amsfonts}
\usepackage{mathtools}
\usepackage{bm}
\usepackage{fancyhdr}
\usepackage{framed}
\usepackage{subfig}
\usepackage{verbatim}
\usepackage{tikz}
\usepackage{relsize}
\usepackage{mathrsfs}
\usepackage{todonotes}
\usepackage{pgfplots}
\usepackage{grffile}
\usepackage{hyperref}

\usetikzlibrary{arrows}

\usepackage{bigints}
\usepackage{todonotes}
\usepackage[font={footnotesize},labelfont=bf]{caption}

\usepackage{setspace}

\newcommand{\bs}{\boldsymbol}

\newcolumntype{C}[1]{>{\centering\let\newline\\\arraybackslash}m{#1}}

\title{\LARGE \bf
A Distributed Predictive Control Approach for Cooperative Manipulation of Multiple Underwater Vehicle Manipulator Systems}

\author{Shahab Heshmati-alamdari,  George C. Karras and Kostas J. Kyriakopoulos% <-this % stops a space
\thanks{The authors are with the Control Systems Lab, Department of Mechanical Engineering, National Technical University of Athens, 9 Heroon Polytechniou Street, Zografou 15780, Greece. Email: {\tt\small \{shahab,karrasg,kkyria@mail.ntua.gr\}} }%
}

\begin{document}
\maketitle \thispagestyle{empty} \pagestyle{empty}

\begin{abstract} 
This paper addresses the problem of cooperative object transportation for multiple Underwater Vehicle Manipulator Systems (UVMSs) in a constrained workspace involving static obstacles. We propose a Nonlinear Model Predictive Control (NMPC) approach for a team of UVMSs in order to transport an object while avoiding significant constraints and limitations such as: kinematic and representation singularities, obstacles within the workspace, joint limits and control input saturations. More precisely, by exploiting the coupled dynamics between the robots and the object, and using certain load sharing coefficients, we design a distributed NMPC for each UVMS in order to cooperatively  transport the object within the workspace's feasible region. Moreover, the control scheme adopts load sharing among the UVMSs according to their specific payload capabilities. Additionally, the feedback relies on each UVMS's locally measurements and no explicit data is exchanged online among the robots, thus reducing the required communication bandwidth. Finally, real-time simulation results conducted in UwSim dynamic simulator running in ROS environment verify the efficiency of the theoretical finding.
\end{abstract}

\section{Introduction}
During the last decades, Unmanned Underwater Vehicles (UUVs) have been widely used in various applications such as marine science (e.g., biology, oceanography, archeology) and offshore industry (e.g., ship maintenance, inspection of oil/gas facilities) \cite{heshmati2018cooperative}. In particular, a vast number of the aforementioned applications, demand the underwater vehicle to be enhanced with intervention capabilities as well \cite{Ridao2015227,Heshmati-alamdari201711197}, thus raising increasing significant scientific interest on Underwater Vehicle Manipulator System (UVMS) lately \cite{Farivarnejad201482,Marani200915,heshmati2018robust}. For instance, some recent European projects: TRIDENT\cite{Fernandez2013121,Simetti2014364,Prats201219,Ribas20152583}, PANDORA \cite{Carrera2014}, and the most recent one DexROV \cite{Gancet2015218}, have boosted significantly the autonomous underwater interaction tasks.

%\begin{figure}[t!]
%	\centering
%	\setlength{\fboxsep}{0pt}%
%	\setlength{\fboxrule}{2pt}%
%	\fbox{\includegraphics[width=0.45\textwidth]{UVMS_coop.jpg}}
%	\caption{Custom-made UVMSs under cooperative transportation.}
%	\label{fig:uvms_frames}
%\end{figure}

Most of the underwater manipulation tasks can be carried out more efficiently, if multiple UVMSs are cooperatively involved. On the other hand, underwater multi-robot tasks are very demanding, with the most significant challenge being imposed by the strict communication constraints \cite{Marani200915,Cui20101491}. Therefor, employing communication based control structure in underwater environment may result in severe performance problems owing to the limited bandwidth and update rate of underwater acoustic devices. Moreover, the number of operating underwater robots in this case, is strictly limited owing to the narrow bandwidth of acoustic communication devices \cite{Stilwell20002358}. To overcome such limitations, recent studies on underwater cooperative manipulation are dealing with designing control schemes under lean communication requirements \cite{Heshmati_AUV2018}.
% Therefore, the design of decentralized cooperative manipulation algorithms for underwater tasks under lean communication becomes apparent.

Cooperative manipulation has been well-studied in the literature, especially the centralized schemes \cite{nikou2017nonlinear}. Despite its efficiency, centralized control is less robust,  and its complexity increases rapidly as the number of participating robots becomes large. On the other hand, decentralized cooperative manipulation schemes usually depend on explicit communication interchange among the robots \cite{verginis2018communication}. For instance, in recent studies \cite{Conti2015261,Furferi20161}, potential fields methods were employed and a multi layer control structure was developed to manage the guidance of UVMSs and the manipulation tasks. Moreover, interesting results have been given in \cite{Simetti2016,Manerikar2015,Manerikar2015523,simetti2015cooperation} where a commonly agreed task space velocity are achieved by transferring data among the robots. However, employing the aforementioned strategies, requires each robot to communicate with the whole robot team, which consequently restricts the number of robots involved in the cooperative manipulation task owing to bandwidth limitations.
\begin{figure}[t!]
	\centering
	\includegraphics[width=1.0\linewidth]{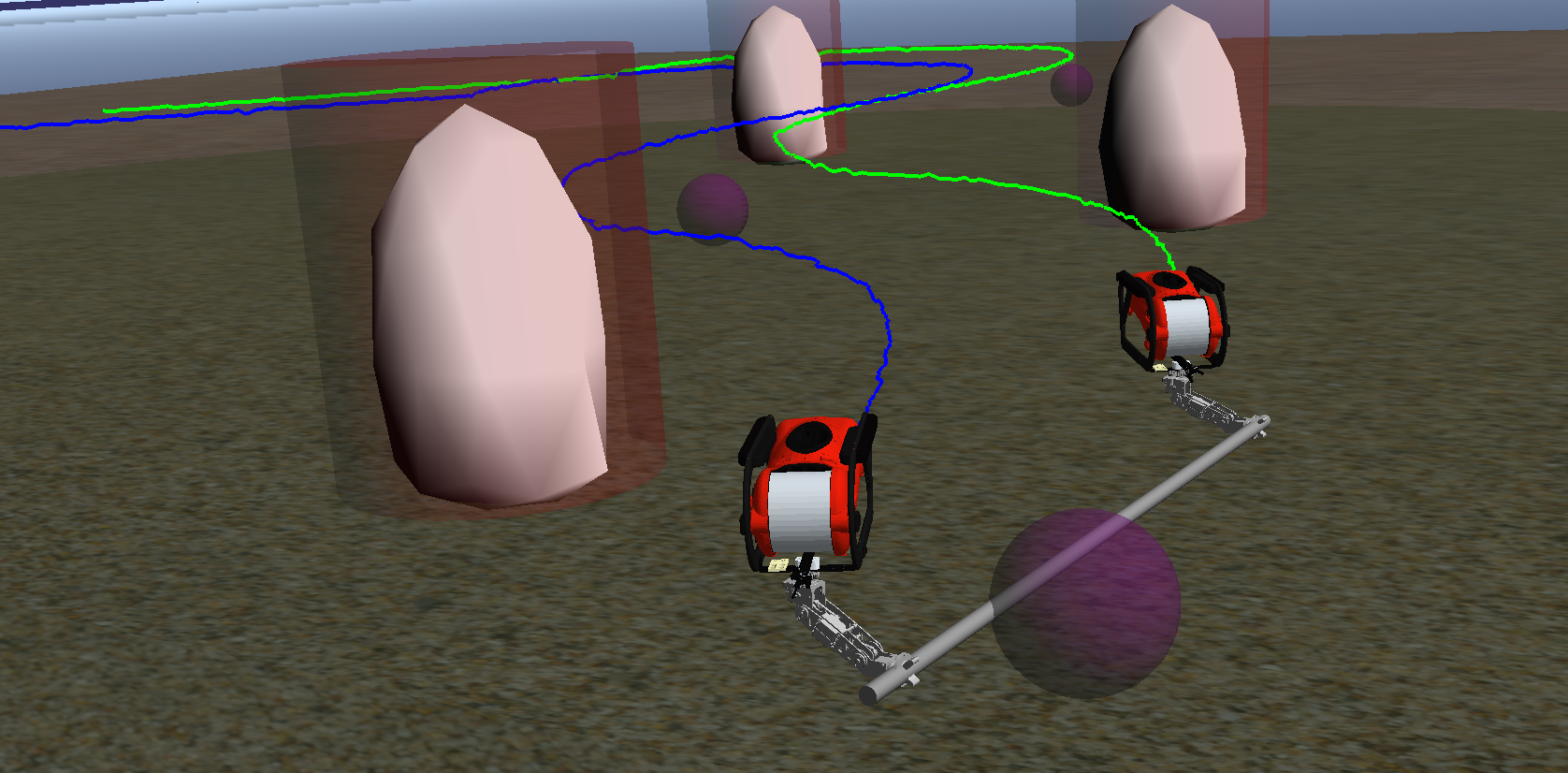}\vspace{0mm}
	\caption{Cooperatively object transportation using two UVMSs inside a constrained workspace including obstacles.}\vspace{-5mm}
	\label{fig:UVMS}
\end{figure}
Moreover, regarding cooperative manipulation, various studies can be found on the literature employing decentralized control schemes where robotic agents use only their local information or observe \cite{Schneider1992383,gudino2004control,caccavale2000task}. Most of the aforementioned studies assume that the robots are equipped with a force/torque sensor on their end effectors in order to acquire knowledge of the interaction contact forces/torques between the end effector and the common object, which may lead to a performance reduction due to sensor noise \cite{farivarnejad2014multiple,moosavian1998multiple,moosavian2010cooperative}. In addition, in most of the studies dealing with cooperative manipulation in literature, very important properties concerning the robotic manipulator systems such as: singular kinematic configurations of Jacobian matrix and joint limits have not been considered at all. 

In this work, the problem of distributed cooperative object transportation considering multiple UVMSs in a constrained workspace with static obstacles is addressed. Specifically, given $N$ UVMSs rigidly grasp a common object, we design distributed controllers for each UVMS in order to navigate the object from an initial position to the final one, while avoiding significant constraints and limitations such as: kinematic and representation singularities, obstacles within the workspace, joint limits and control input saturations. More precisely, by exploiting the coupled dynamics between the robots and the object and by using certain load sharing coefficients we design a distributed Nonlinear Model Predictive Control (NMPC) \cite{allgower2004} for each UVMS in order to transport cooperatively the object and steer it along of a computed feasible path within the workspace. The design of that feasible path is based on the Navigation Function concept \cite{Koditschek1990412} which is adopted here in order to achieve distributed consensus on the object's desired trajectory as well to avoid collisions with the obstacles and the workspace boundary.  In proposed control strategy we also take into account constraints that emanate from control input saturation as well kinematic and representation singularities.  Moreover, the control scheme adopts load sharing among the UVMSs according to their specific payload capabilities. Finally, the feedback relies on each UVMS's locally measurements i.e., position and velocity measurements (e.g., sensor fusion based on measurement of various onboard sensors such as IMU, USBL and DVL) and no explicit data is exchanged online among the robots. This, consequently, increases significantly the robustness of the cooperative scheme and furthermore avoids any restrictions imposed by the acoustic communication bandwidth (e.g., the number of participating UVMSs).

%One should bear in mind that although underwater vehicles are equipped with acoustic modems for communication with the surface control station or with other robots, employing implicit communication based cooperative control protocols is clearly motivated by the limited bandwidth of acoustic communication devices. Moreover, in order to achieve collision avoidance, either the leader has to transmit online the desired object trajectory to the followers, or all UVMSs should obtain a mutually agreed desired trajectory of the object, which necessitates for an accurate common localization system \cite{Conti2015261} that is extremely challenging and prone to errors in underwater environments. On the contrary, in the proposed scheme, it is worth noting that each follower estimates locally and in a distributed way, the desired object trajectory relatively to its inertial frame, employing exclusively its own measurements (position, velocity and force/toque). In this way, although the proposed control strategy does not remove all practical needs for communication in underwater intervention tasks, nevertheless, it relieves the team of robots from intense inter-robot communication during the execution of the collaborative tasks. This, consequently, increases significantly the robustness of the cooperative scheme and furthermore avoids any restrictions imposed by the acoustic communication bandwidth (e.g., the number of participating UVMSs). \vspace{-0.2cm}

\section{Mathematical Modeling}
Consider $N$ UVMSs rigidly grasping an object within a constrained workspace with static obstacles (see Fig \ref{fig:UVMS}). We assume that each UVMS is fully-actuated at its end-effector frame. We also assume that the UVMSs are equipped with appropriate sensors, that allow them to measure their position  and velocity. 
%Additionally, the geometric parameters of the both UVMSs and the commonly grasped object are considered known. Moreover, the control of each UVMS will be designed based on a commonly agreed frame on a specific feature of the object, which could be identified employing a visual detection system, owing to the fact that the limited underwater visibility is not an issue when all robots are close to the object of interest. 
\subsection{UVMS Kinematics}
  Consider $N$ UVMSs operating in a bounded workspace $\mathcal{W} \subseteq \mathbb{R}^3$. First, we denote the coordinates of each UVMS's end effector by $\bs{p}_i=[\bs{\eta}_{1,p_i}^\top,\bs{\eta}_{2,p_i}^\top]^\top$ where $\bs{\eta}^\top_{1,p_i}=[x_{p_i},y_{p_i},z_{p_i}]^\top$ and $\bs{\eta}^\top_{2,p_i}=[\phi_{p_i},\theta_{p_i},\psi_{p_i}]^\top$ denote the position and the orientation expressed in Euler angles representation with respect to (w.r.t) the inertial frame.  Let $\bs{q}_i=[\bs{q}^\top_{B,i},~\bs{q}^\top_{m,i}]^\top\in \mathbb{R}^{n_i}$, with $n_i\in \mathbb{N}$, $i\in\mathcal{N}$  be the joint state variables of each UVMS,  where $\bs{q}_{B,i}=[\bs{\eta}^\top_{1,{B_i}},\bs{\eta}^\top_{2,{B_i}}]^\top$ is the vector that involves the position $\bs{\eta}^\top_{1,B_i}$ and the orientation $\bs{\eta}^\top_{2,B_i}$ of the vehicle and $\bs{q}_{m,i}$ is the vector of the angular positions of the manipulator's joints.  Specifically, $\bs{\eta}^\top_{1,B_i}=[x_{B_i},y_{B_i},z_{B_i}]^\top$ and $\bs{\eta}^\top_{2,B_i}=[\phi_{B_i},\theta_{B_i},\psi_{B_i}]^\top,~ i\in \{O,1,\ldots,N\}$ denote the position and the orientation expressed in Euler angles representation w.r.t the inertial frame. Let also define the UVMS' end effector generalized velocities by ${\bs{v}}_i=[\dot{\bs{\eta}}_{1,i}^\top,\bs{\omega}_i^\top]^\top,~ i\in \mathcal{N}$, where $\dot{\bs{\eta}}_{1,i}$ and $\bs{\omega}_i$ denote the linear and angular velocity respectively. In addition, the position and orientation of the UVMS
  end-effector w.r.t inertial frame, is given by the forward  kinematics of the complete system (arm and vehicle base) as
  follows:\vspace{-2mm}
    \begingroup\makeatletter\def\f@size{9.5}\check@mathfonts
  \def\maketag@@@#1{\hbox{\m@th\large\normalfont#1}}\begin{align}
  \bs{p}_i=\mathcal{F}(\bs{q}_i)~,i\in\mathcal{N}\label{FWD_kin}
  \end{align}\endgroup  \vspace{-6mm}
  
\noindent Moreover, for the augmented UVMS system we have \cite{antonelli}:\vspace{-1mm}
   \begingroup\makeatletter\def\f@size{9.5}\check@mathfonts
 \def\maketag@@@#1{\hbox{\m@th\large\normalfont#1}} \begin{equation}
 {\bs{v}}_i= \bs{J}_i({\bs{q}}_i)\dot{\boldsymbol{{{q}}}}_i,~ i\in\mathcal{N}\label{eq222}
  \vspace{-1mm}\end{equation}\endgroup
 where $\dot{\boldsymbol{q}}_i=[  \dot{\bs{q}}_{B,i}^\top,\dot{\bs{q}}_{m,i}^\top]^\top \in \mathbb{R}^{n_i}$  is the velocity vector involving the velocities of the vehicle w.r.t the inertial frame as well as the joint velocities of the manipulator and $ \bs{J}_i(\bs{q}_i)$ is the geometric Jacobian matrix \cite{antonelli}. Note that the $\bs{J}_i({\bs{q}}_i)$ becomes singular at kinematic singularities defined by the set \vspace{-2mm}
   \begingroup\makeatletter\def\f@size{9.5}\check@mathfonts
 \def\maketag@@@#1{\hbox{\m@th\large\normalfont#1}}\begin{align}
 Q_{s_i}=\{\bs{q}_i\in\mathbb{R}^{n_i}:\det(\bs{J}_i({\bs{q}}_i)[\bs{J}_i({\bs{q}}_i)]^\top\!)=0\},~i\in\mathcal{N}.\!\!\label{singular_robot}
\vspace{-2mm} \end{align}\endgroup

\subsection{UVMS Dynamics}
The dynamics of a UVMS after straightforward algebraic manipulations can be written as \cite{antonelli}:\vspace{-1mm}
  \begingroup\makeatletter\def\f@size{9.5}\check@mathfonts
\def\maketag@@@#1{\hbox{\m@th\large\normalfont#1}}\begin{align}
\!\!\!\!\bs{{M}}\!_{q_i}\!(\bs{q}_i)\ddot{\bs{q}}_i\!+\!\bs{{C}}\!_{q_i}\!(\dot{\bs{q}}_i,\bs{q}_i)\dot{\bs{q}}_i\!+\!\bs{{D}}\!_{q_i}\!(\dot{\bs{q}}_i,\!\bs{q}_i)\dot{\bs{q}}_i\!+\!\bs{{g}}_{q_i}\!&(\!\bs{q}_i\!)\!=\boldsymbol{\tau}_i\!\!-\!{\bs{J}_i}\!^\top\!\!\boldsymbol{\lambda}_i\!\!\label{eq6}
\vspace{-2mm}\end{align}\endgroup
for $i\in\mathcal{N}$, where $\boldsymbol{\lambda}_i$ is the vector of generalized interaction forces and torques that UVMS exerts on the object, $\boldsymbol{\tau}_i$ denotes the vector of control inputs
(forces and torques), $\bs{{M}}_{q_i}(\bs{q}_i)$ is the inertial matrix, $\bs{{C}}_{q_i}(\dot{\bs{q}}_i,\bs{q}_i)$ represents coriolis and centrifugal terms, $\bs{{D}}_{q_i}(\dot{\bs{q}}_i,\bs{q}_i)$ models dissipative effects and $\bs{{g}}_i(\bs{q}_i)$ encapsulates the gravity and buoyancy effects. In view of \eqref{eq222} we have:\vspace{-2mm}
  \begingroup\makeatletter\def\f@size{9.5}\check@mathfonts
\def\maketag@@@#1{\hbox{\m@th\large\normalfont#1}}\begin{align}
\dot{\bs{v}}_i=\bs{J}_i(\bs{q}_i)\ddot{\bs{q}}_i+\dot{\bs{J}}_i(\bs{q}_i)\dot{\bs{q}}_i,~i\in\mathcal{N}\label{der_upsilon}
\end{align} \endgroup\vspace{-5mm}

\noindent where $\dot{\bs{J}}_i(\bs{q}_i)\in \mathbb{R}^{6\times n_i}$ represents the Jacobian derivative function. Then, by employing the differential kinematics \eqref{eq222} as well as \eqref{der_upsilon}, we obtain from \eqref{eq6} the transformed task space dynamics \cite{Siciliano-b1129198}:\vspace{-1mm}
  \begingroup\makeatletter\def\f@size{9.5}\check@mathfonts
\def\maketag@@@#1{\hbox{\m@th\large\normalfont#1}}\begin{align}
\!\bs{{M}}\!_i(\bs{q}_i)\dot{\bs{v}}_i\!+\!\bs{{C}}\!_{i}(\dot{\bs{q}}_i,\!\bs{q}_i){\bs{v}}_i\!+\!\bs{{D}}_{i}(\dot{\bs{q}}_i,\bs{q}_i){\bs{v}}_i\!+\!\bs{{g}}_{i}(\bs{q}_i)\!=\!\boldsymbol{u}_i\!\!-\!\!\boldsymbol{\lambda}_i\!\!\label{UVMS_task_space}
\end{align}\endgroup\vspace{-5mm}

\noindent for all $i\in \mathcal{N}$ with  corresponding task space terms $\bs{M}_{i}\in \mathbb{R}^{6\times 6}$, $\bs{C}_{i}\in \mathbb{R}^{6\times 6}$, $\bs{D}_{i}\in \mathbb{R}^{6\times 6}$, $\bs{g}_{i}\in \mathbb{R}^{6}$ with $\bs{u}_i\in \mathbb{R}^{6}$ to be the vector of task space generalized forces/torques. The task space dynamics \eqref{UVMS_task_space} can be written in vector form as:   
\begin{align}
\bs{{M}}(\bs{q})\dot{\bs{v}}+\bs{{C}}(\dot{\bs{q}},\bs{q}){\bs{v}}+\bs{{D}}(\dot{\bs{q}},\bs{q}){\bs{v}}+\bs{{g}}(\bs{q})=\boldsymbol{u}-\boldsymbol{\lambda}\label{UVMS_Overal_task_space}
\end{align}
  \begingroup\makeatletter\def\f@size{9.5}\check@mathfonts
\def\maketag@@@#1{\hbox{\m@th\large\normalfont#1}}where $\bs{v}=[\bs{v}_1^\top,\ldots,\bs{v}_N^\top]^\top\in \mathbb{R}^{6N}$, $\bs{M}=\text{diag}\{[\bs{M}_i]\}\in \mathbb{R}^{6N\times6N}$, $\bs{C}=\text{diag}\{[\bs{C}_i]\}\in \mathbb{R}^{6N\times6N}$, $\bs{D}=\text{diag}\{[\bs{D}_i]\}\in \mathbb{R}^{6N\times6N}$, $\bs{\lambda}=[\bs{\lambda}_1^\top,\ldots,\bs{\lambda}_{N}^\top]^\top$, $\bs{u}=[\bs{u}_1^\top,\ldots,\bs{u}_{N}^\top]^\top$, $\bs{d}=[\bs{d}_1^\top,\ldots,\bs{d}_{N}^\top]^\top$, $\bs{g}=[\bs{g}_1^\top,\ldots,\bs{g}_{N}^\top]^\top \in \mathbb{R}^{6N}$.\endgroup

\subsection{Object Dynamic}
We denote the object's coordinate and its generalized velocities by $\bs{x}_O=[\bs{\eta}_{1,O}^\top,\bs{\eta}_{2,O}^\top]^\top$  and ${\bs{v}}_O=[\dot{\bs{\eta}}_{1,O}^\top,\bs{\omega}_O^\top]^\top$ respectively, with $\bs{\eta}^\top_{1,O}=[x_{O},y_{O},z_{O}]^\top$ and $\bs{\eta}^\top_{2,O}=[\phi_{O},\theta_{O},\psi_{O}]^\top\!\!$. The dynamics of the object can be given \cite{antonelli}:  \vspace{-4mm}
  \begingroup\makeatletter\def\f@size{9.5}\check@mathfonts
\def\maketag@@@#1{\hbox{\m@th\large\normalfont#1}}\begin{subequations}\label{obj_dyn}
\begin{align}
&\!\!\!\!\!\dot{\bs{x}}_O={\bs{J}'_O({\bs{\eta}}_{2,O})}^{-1}\bs{v}_O\label{obj_dyn_a}\\
&\!\!\!\!\!\bs{M}\!_O(\bs{x}_O)\dot{\bs{v}}_O\!\!+\!\bs{C}\!_O(\bs{v}_O,\!\bs{x}_O)\bs{v}_O\!\!+\!\!\bs{D}\!_O(\bs{v}_O,\!\bs{x}_O\!)\bs{v}\!_O\!\!+\!\!\bs{g}_O\!\!=\!\!\bs{\lambda}_O\!\!\!\!\label{obj_dyn_b}\end{align}
\end{subequations}\endgroup
where $\bs{M}_O(\bs{x}_O)$ is the positive definite inertia matrix, $\bs{C}_O(\bs{v}_O,\bs{x}_O)$ is the Coriolis matrix, $\bs{g}_O$ is the vector of gravity and buoyancy effects, $\bs{D}_O(\bs{v}_O,\bs{x}_O)$ models dissipative effects and $\bs{\lambda}_O$ is the vector of generalized forces acting on the object's center of mass. Moreover, ${\bs{J}'_O({\bs{\eta}}_{2,O})}$ is the object representation Jacobian that transforms the Euler angle rates into velocity $\bs{\omega}_O$ and is singular when when $\theta_{O}=\pm \frac{\pi}{2}$\cite{antonelli}. 
\section{Control Methodology}
%We assume that the all of UVMS are aware of both the desired configuration of the object as well as of the obstacles position in the workspace\footnote{The desired configuration of the object can be transmitted to each UVMS before executing the cooperation task.} The proposed approach builds on designing a NMPC scheme for the system of the UVMSs and the object.  
First, the overall dynamics of the system are formulated which are decoupled next among the object and the robots by using certain load
sharing coefficients. Each UVMS at each sampling time, solves a NMPC subject to its corresponding part of that overall dynamics and a number of inequality constraints that incorporate its internal limitations (e.g., joint limits, kinematic and representation singularities, collision between the arm and the base, manipulability) in order to drive cooperatively the object and steer it along of a computed feasible path within the workspace. The computation of that feasible path is based on the concept of Navigation Functions \cite{Koditschek1990412} that is incorporated to deal with consensus on a mutually agreed trajectory of the commonly object. \vspace{-2mm}
\begin{comment}Moreover, we assume that each UVMS $i\in\mathcal{N}$ is able to continuously measure its state vector $\bs{q}_i, \dot{\bs{q}}_i ~i\in \mathcal{N}$ based on its own state measurements (sensor fusion of locally onboard navigation system sensors, e.g., DVL, IMU, USBL).\end{comment}

\subsection{Coupled Dynamics}
Owing to the rigid grasp of the object, the following equations hold:  \vspace{-3mm}
 \begingroup\makeatletter\def\f@size{9.5}\check@mathfonts
 \def\maketag@@@#1{\hbox{\m@th\large\normalfont#1}}\begin{align}
 \bs{p}_i=\bs{x}_O+ \left[
 \begin{array}{ccc}
 ^I\bs{R}_O \bs{l}_i\\
 \boldsymbol{\alpha}_i\end{array} \right],~ i\in\mathcal{N} \label{eq3}
 \end{align}\endgroup  \vspace{-3mm}
 
\noindent where the vectors $\bs{l}_i=[l_{ix},l_{iy},l_{iz}]^\top$ and $\boldsymbol{\alpha}_i=[\alpha_{ix},\alpha_{iy},\alpha_{iz}]^\top,~ i\in\mathcal{N} $ represent the \emph{constant} relative position and orientation of the end-effector w.r.t the object, expressed in the object's frame and  $^I\bs{R}_O$ denotes the rotation matrix between the object and the inertial frame $\{I\}$. Thus, using \eqref{eq3} each UVMS can compute the object's position w.r.t inertial frame $\{I\}$, since the object geometric parameters are considered known. Furthermore,  due to the grasping rigidly, it holds that $\bs{\omega}_i=\bs{\omega}_O,~i\in \mathcal{N}$, one obtains:  \vspace{-2mm}
  \begingroup\makeatletter\def\f@size{9.5}\check@mathfonts
 \def\maketag@@@#1{\hbox{\m@th\large\normalfont#1}}\begin{align}
 {\bs{v}}_O=\bs{J}_{i_O}{\bs{v}}_i,~ i\in\mathcal{N}\label{eq4}
 \end{align}\endgroup  \vspace{-5mm}
 
 \noindent where $\bs{J}_{i_O},~i \in\mathcal{N}$ denotes the Jacobian from the end-effector of each UVMS to the object's center of mass, that is defined as: 
  \begingroup\makeatletter\def\f@size{9.5}\check@mathfonts
 \def\maketag@@@#1{\hbox{\m@th\large\normalfont#1}}\begin{align*}
 \bs{J}_{i_O}=\arraycolsep=5.4pt \left[
 \begin{array}{ccc}
 \bs{I}_{3\times 3} & -\bs{S}(\bs{l}_i)\\
 \bs{0}_{3\times 3} & \bs{I}_{3\times 3}
 \end{array} \right]\in \mathbb{R}^{6\times 6},~~i \in\mathcal{N}%\label{J_oi}
 \end{align*}\endgroup
 where $\bs{S}(\bs{l}_i)$ is the skew-symmetric matrix of vector $\bs{l}_i=[l_{ix},l_{iy},l_{iz}]^\top$. Notice that $\bs{J}_{i_O},~i\in\mathcal{N}$ are always full-rank owing to the grasp rigidity and hence obtain a well defined inverse. Thus, the object's velocity can be easily computed via \eqref{eq4}. Moreover, from \eqref{eq4}, one obtains the acceleration relation:\vspace{-3mm}
  \begingroup\makeatletter\def\f@size{9.5}\check@mathfonts
 \def\maketag@@@#1{\hbox{\m@th\large\normalfont#1}}\begin{align}
 \dot{\bs{v}}_O=\bs{J}_{i_O}\dot{\bs{v}}_i+\dot{\bs{J}}_{i_O}\bs{v}_i,~i\in\mathcal{N}\label{dotv_obj}
 \end{align}\endgroup  \vspace{-4mm}
 
 \noindent which will be used in the subsequent analysis. In addition, the kineto-statics duality along with the grasp rigidity suggest that the force $\bs{\lambda}_O$ acting on the object's center of mass and the generalized forces $\bs{\lambda}_i,~i\in \mathcal{N}$, exerted by the UVMSs at the grasping points, are related through: \vspace{-2mm}
   \begingroup\makeatletter\def\f@size{9.5}\check@mathfonts
 \def\maketag@@@#1{\hbox{\m@th\large\normalfont#1}}\begin{align}\label{object_grasp_matrix}
 \bs{\lambda}_O=\bs{G}^\top\bs{\lambda}
 \end{align}\endgroup  \vspace{-7mm}
\noindent where:
  \begingroup\makeatletter\def\f@size{9.5}\check@mathfonts
\def\maketag@@@#1{\hbox{\m@th\large\normalfont#1}} \begin{align}\label{grasp matrix}
 \bs{G}=\Big[[\bs{J}_{O_1}]^\top,\ldots,[\bs{J}_{O_N}]^\top\Big]^\top\in \mathbb{R}^{6N\times 6 }
 \end{align}\endgroup
 is the full column-rank grasp matrix, $\bs{J}_{O_i}=[\bs{J}_{i_O}]^{-1},~i\in\mathcal{N}$  and $\bs{\lambda}=[\bs{\lambda_1}^\top,\ldots,\bs{\lambda_{N}}^\top]^\top$ is the vector of overall interaction forces and torques. 
% \begin{remark}\label{inertial dynamics}
% 	Wrenches that lie on the null space of the grasp matrix $\bs{G}^\top$ do not contribute to the object dynamics. Therefore, we may incorporate in the control scheme an extra component $\bs{\lambda}_{int,i}=\!(\bs{I}\!\!-\! (\bs{G}^\top\!)^\#\!\bs{G}^\top\!)\bs{\lambda}^d_{int}$, $i\in \mathcal{N}$, that belongs to the null space of $\bs{G}^\top$, in order to regulate the steady state internal forces, where $(\bs{G}^\top\!)^\#$ denotes the generalized inverse of $\bs{G}^\top$. Notice that owing to the rigid grasp, $l_i,~i\in \mathcal{N}$ remain constant. Thus, since $l_i,~i\in \mathcal{N}$ are considered known to the team of UVMSs\footnote{This can be achieved by using the acoustic modems before beginning the task execution.}, if  $\bs{\lambda}^d_{int}$ is chosen constant, no communication is needed during task execution in order to compute $\bs{G}^\top,~(\bs{G}^\top\!)^\#$  and $\bs{\lambda}_{int,i}$.
% \end{remark}
By substituting \eqref{UVMS_Overal_task_space} into \eqref{object_grasp_matrix} one obtains:  \vspace{-4mm}
  \begingroup\makeatletter\def\f@size{9.5}\check@mathfonts
\def\maketag@@@#1{\hbox{\m@th\large\normalfont#1}}\begin{align}
\bs{\lambda}=\bs{G}^\top\Big[\boldsymbol{u}-\bs{{M}}(\bs{q})\dot{\bs{v}}-\bs{{C}}(\dot{\bs{q}},\bs{q}){\bs{v}}-\bs{{D}}(\dot{\bs{q}},\bs{q}){\bs{v}}-\bs{{g}}(\bs{q})\Big]
\end{align}\endgroup
which, after substituting  \eqref{eq4}, \eqref{dotv_obj}, \eqref{obj_dyn} and rearranging terms, yields the overall system coupled dynamics:
  \begingroup\makeatletter\def\f@size{9.5}\check@mathfonts
\def\maketag@@@#1{\hbox{\m@th\large\normalfont#1}}\begin{align}
\widetilde{\bs{M}}(\tilde{\bs{q}}_{ov})\dot{\bs{v}}_O\!+\!\widetilde{\bs{C}}(\tilde{\bs{q}}_{ov}){\bs{v}}_O\!+\!\widetilde{\bs{D}}(\tilde{\bs{q}}_{ov}){\bs{v}}_O\!+\!\widetilde{\bs{g}}(\tilde{\bs{q}}_{ov})\!=\!\bs{G}^\top \bs{u}\label{Overall_dynamics}
\end{align}\endgroup
where $\tilde{\bs{q}}_{ov}=[\bs{q}^\top\!\!\!,\dot{\bs{q}}^\top\!\!\!,\bs{x}_O^\top,{\bs{v}}_O^\top]^\top$ and:
\begingroup\makeatletter\def\f@size{8.6}\check@mathfonts
\def\maketag@@@#1{\hbox{\m@th\large\normalfont#1}}\begin{gather*}
\widetilde{\bs{M}}(\tilde{\bs{q}}_{ov})=\bs{M}_O(\bs{x}_O)+\bs{G}^\top\bs{M}(\bs{q})\bs{G}\\
\!\widetilde{\bs{C}}(\tilde{\bs{q}}_{ov})\!=\!\bs{C}_O(\bs{v}_O,\bs{x}_O)\!+ \! \bs{G}^\top\bs{M}(\bs{q})\dot{\bs{G}}(\dot{\bs{q}},\bs{q})     \! +\!\bs{G}^\top\bs{C}(\dot{\bs{q}},\bs{q})\bs{G}\\
\!\widetilde{\bs{D}}(\tilde{\bs{q}}_{ov})\!=\!\bs{D}_O(\bs{v}_O,\bs{x}_O)\!+\!\bs{G}^\top\!\!\bs{C}(\dot{\bs{q}},\bs{q})\bs{G},~\widetilde{\bs{g}}(\tilde{\bs{q}}_{ov})\!=\!\bs{g}_O(\bs{x}_O)\!+\!\bs{G}^\top\!\!\bs{g}(\bs{q})
\end{gather*}\endgroup
Now, consider the design constants $c_i,~i\in\mathcal{N}$ satisfying: \vspace{-2mm}
  \begingroup\makeatletter\def\f@size{9.5}\check@mathfonts
\def\maketag@@@#1{\hbox{\m@th\large\normalfont#1}}\begin{align}\label{c_i}
c_i\in (0,1),\forall i\in\mathcal{N}\qquad ~\text{and}~\qquad \sum_{i\in\mathcal{N}} c_i=1,~ 
\end{align}\endgroup   \vspace{-4mm}

\noindent that we introduce here in order to act as the load sharing coefficients for the team of UVMS. In view of \eqref{c_i}, by employing \eqref{object_grasp_matrix}, \eqref{eq222}, \eqref{der_upsilon}, \eqref{eq4} and \eqref{dotv_obj}, and after straightforward algebraic
manipulations, the overall coupled dynamics of \eqref{Overall_dynamics} can be divided and rewritten as: 
  \begingroup\makeatletter\def\f@size{9.4}\check@mathfonts
\def\maketag@@@#1{\hbox{\m@th\large\normalfont#1}}\begin{align}
\sum_{i\in\mathcal{N}}\!\!\!\Big\{ \widetilde{\bs{M}}_i({\bs{q}}_i)\ddot{\bs{q}}_i\!+\!\widetilde{\bs{C}}_i(\dot{\bs{q}}_i,{\bs{q}}_{i}){\dot{\bs{q}}}_i\!+\!\widetilde{\bs{D}}_i(\dot{\bs{q}}_i,{\bs{q}}_{i}){\dot{\bs{q}}}_i\!+\!\widetilde{\bs{g}}_i({\bs{q}}_{i})\Big\}\!\!=\!\!\!\sum_{i\in\mathcal{N}}\!\!\!\bs{J}\!_{O_i}^\top\!\bs{u}_i\!\!\label{dec_dynamics}
\end{align}\endgroup   \vspace{-8mm}

\noindent  where: \vspace{-8mm}

\begingroup\makeatletter\def\f@size{8.6}\check@mathfonts
\def\maketag@@@#1{\hbox{\m@th\large\normalfont#1}}\begin{gather*}
\widetilde{\bs{M}}_i({\bs{q}}_i)=c_i\bs{M}_O\bs{J}_{i_O}\bs{J}_i+\bs{J}_{O_i}^\top\bs{M}_i\bs{J}_i\\
\widetilde{\bs{C}}_i(\dot{\bs{q}}\!_i,\!{\bs{q}}_{i})\!\!=\!c_i\!\Big[\!\bs{M}\!_O\bs{J}\!_{i_O}\!\dot{\bs{J}}\!_i\!\!+\!\bs{M}\!_O\dot{\bs{J}}\!_{i_O}\bs{J}\!_i\!\!+\!\bs{C}\!_O\!\bs{J}\!_{i_O}\!\bs{J}\!_i\!\Big]\!\!+\!\bs{J}\!_{O_i}^\top\!\Big[\!\bs{M}\!_i\dot{\bs{J}}\!_i\!+\!\bs{C}\!_i\bs{J}\!_i\! \Big]\\
\widetilde{\bs{D}}_i(\dot{\bs{q}}_i,{\bs{q}}_{i})=c_i\bs{D}_O\bs{J}_{i_O}\bs{J}_i\!+\!\bs{J}_{O_i}^\top\bs{D}_i\bs{J}_i,\quad\widetilde{\bs{g}}_i({\bs{q}}_{i})=c_i\bs{g}_O+\bs{J}_{O_i}^\top\bs{g}_i
\end{gather*}\endgroup
which is the distributed version of \eqref{Overall_dynamics}, since for each UVMS, it is based only individually on its locally measurements (i.e., $\bs{q}_i$ and $\dot{\bs{q}}_i$). Now, by using the notation $\bs{x}_i=[\bs{q}_i^\top,\dot{\bs{q}}_i^\top]^\top$, the decentralized dynamics of each UVMS based on \eqref{dec_dynamics}, can be written as compact form:\vspace{-0mm}
  \begingroup\makeatletter\def\f@size{9.5}\check@mathfonts
\def\maketag@@@#1{\hbox{\m@th\large\normalfont#1}}\begin{align}
\dot{\bs{x}}_i=f_i(\bs{x}_i,\bs{u}_i)=\begin{bmatrix}
f_{i_1}(\bs{x}_i)\\
f_{i_2}(\bs{x}_i,\bs{u}_i)
\end{bmatrix}~,i\in\mathcal{N}\label{nominal_system}
\end{align} \endgroup \vspace{-2mm}

\noindent where:\vspace{-4mm}
\begingroup\makeatletter\def\f@size{8.5}\check@mathfonts
\def\maketag@@@#1{\hbox{\m@th\large\normalfont#1}}\begin{gather*}
f_{i_1}(\bs{x}_i)=\dot{\bs{q}}_i\\
f_{i_2}(\bs{x}_i,\bs{u}_i)\!=\!\widetilde{\bs{M}}^\#_i\!({\bs{q}}_i)\Big(\! \bs{J}_{O_i}^\top\!(\bs{q}_i) \bs{u}_i\!\!-\! \widetilde{\bs{C}}_i(\dot{\bs{q}}_i,{\bs{q}}_{i})\dot{\bs{q}}_i\!\!-\!\widetilde{\bs{D}}_i(\dot{\bs{q}}_i,{\bs{q}}_{i})\dot{\bs{q}}_i\!-\!\widetilde{\bs{g}}_i({\bs{q}}_{i}) \! \Big)
\end{gather*}\endgroup
with: \vspace{-6.5mm}
  \begingroup\makeatletter\def\f@size{9.5}\check@mathfonts
\def\maketag@@@#1{\hbox{\m@th\large\normalfont#1}}\begin{align*}
\widetilde{\bs{M}}^\#_i\!({\bs{q}}_i)=   \widetilde{\bs{M}}_i({\bs{q}}_i)\Big[ \widetilde{\bs{M}}_i({\bs{q}}_i)\widetilde{\bs{M}}^\top_i\!\!(\bs{q}_i) \Big]^{-1}
\end{align*}\endgroup
%Moreover, we can define the actual decentralized dynamics of each UVMS based on \eqref{dec_dynamics} and the expression of the nominal system \eqref{nominal_system} as:
%\begin{align}
%\dot{\bs{x}}_i=f_i(\bs{x}_i,\bs{u}_i)+\bs{w}_i(\bs{x}_i)~,i\in\mathcal{N}\label{actual_dynamics}
%\end{align}
%with:
%\begin{align*}
%\bs{w}_i(\bs{x}_i)=\widetilde{\bs{M}}^\#_i\!({\bs{q}}_i)\widetilde{\bs{d}}_i(\dot{\bs{q}}_i,{\bs{q}}_{i},t)\in W_i.
%\end{align*}
%where $W$ is the compact set of external disturbances and uncertainties bounded by $|\bs{w}_i|\leq \bar{w}_i$. 
%\begin{remark}
%Notice that Eq. \eqref{actual_dynamics} is the actual decentralized dynamical equation for UVMS $i\in\mathcal{N}$, since it contains the vector of disturbance effecting on the system. In this way, we consider Eq.\eqref{nominal_system} as the nominal model of UVMS $i\in\mathcal{N}$, in which no disturbances are considered.
%\end{remark}

%\begin{Property}
%	It can be concluded that the nominal system of \eqref{nominal_system} is \emph{Lipschitz continuous} in $X_i$ since it is continuously differentiable in its domain. Thus, for every $\bs{x}_i,\bs{x}'_i,~\in X_i$, with $\bs{x}_i\neq\bs{x}'_i$ there is a positive Lipschitz constant $L_f < \infty$, such that $\|f(\bs{x}_i,\bs{u})\!-\!f(\bs{x}'_i,\!\bs{u})\|\! \le\! L_f \|\bs{x}_i - \bs{x}'_i\|$.
%\end{Property}

\vspace{-4mm}

\subsection{Description of the Workspace}
\begin{figure}[ht!]
	\centering
	\includegraphics[width=0.8\linewidth]{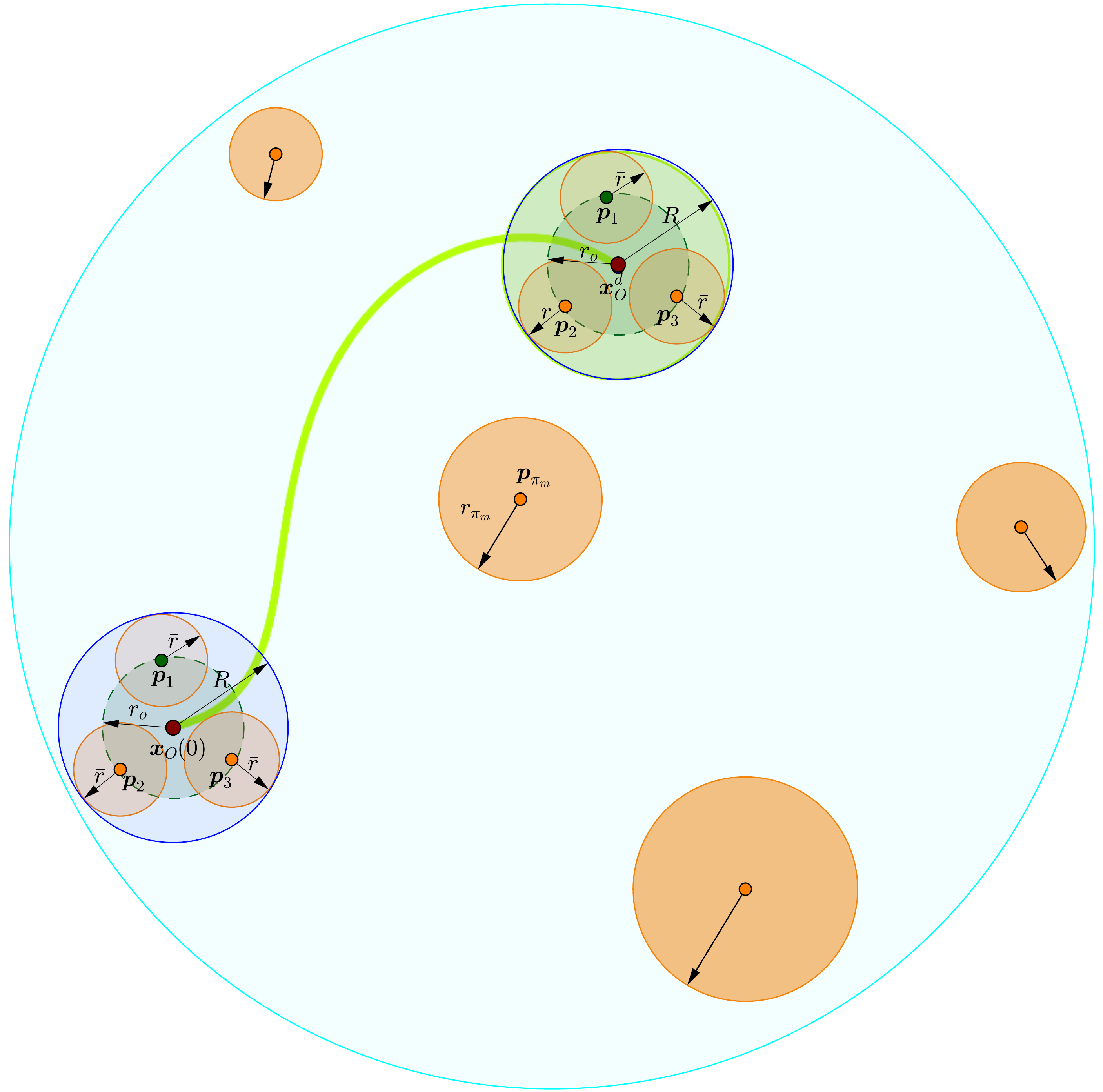}
	\caption{Graphical representation of a safe trajectory of the robotic team. The orange areas indicate the obstacles. The blue line encircles the area covered by the robotic team and the object.} 
	\label{fig:valid_transition}
\end{figure}

In this work, the obstacles, the robots as well as the workspace are all modeled by spheres (i.e., we adopt the spherical world representation \cite{Koditschek1990412}). In this spirit, let $\mathcal{B}(\bs{x}_O, r_0)$ be a closed ball that covers the volume of the object and has radius $r_0$. We also define the closed balls $\mathcal{B}(\boldsymbol{p}_i, \bar{r}), i\in \mathcal{K}$, centered at the end-effector of each UVMS that cover the robot volume for all possible configurations. We also assume that the distance among the grasping points on the given object is at least $2\bar{r}$ \footnote{The distance $2\bar{r}$ denotes the minimum allowed distance at which two bounding spheres $\mathcal{B}(\boldsymbol{p}_i, \bar{r})$  and $\mathcal{B}(\boldsymbol{p}_j, \bar{r})$ $i,j\in \mathcal{K}, i \neq j$ do not collide.}.
%\begin{figure}[ht!]
%	\centering
%	\begin{tikzpicture}[scale = 0.5]	
%	
%	%draw region \pi_k
%	
%	\draw [color = black, fill = blue!20] (-4.0,0) circle (2.5cm);
%	%	\node at (-8.2, 3.7) {$\pi_k$};
%	\draw [color=black,thick,->,>=stealth'](-4.0, 0) to (-4.8, -2.4);
%	
%	\node at (-4.2, -1.6) {$\bar{r}$};
%	
%	
%	\node at (-3.7, -0.6) {$\bs{p}_i$};
%	
%	%draw region \pi_{k'}
%	
%	\draw [color = red, fill = red!20] (1.0, 0) circle (2.5cm);
%	%	\node at (1.0, 3.7) {$\pi_{k'}$};
%	\draw [color=black,thick,->,>=stealth'](1.0, 0) to (2.5, -2.0);
%	
%	\node at (2.2,-1) {$\bar{r}$};
%	%	\node at (4.5, -1.7) {$r_{\pi_{k'}}$};
%	\draw[orange, rotate=0,  line width=3pt] (1,0.05) -- (-4.0,0.05);
%	\node at (-4.0, 0) {$\bullet$};
%	\node at (1.0, 0) {$\bullet$};
%	
%	\node at (0.8, -0.6) {$\bs{p}_j$};
%	%plot trajectory of end-effecotr
%	%	\draw [line width = 0.35mm] (-6.7,-3.0)  ;	
%	\end{tikzpicture}
%	\caption{ Graphical representation of the minimum allowed distance $\bar{r}$. }
%	\label{fig:valid_grasp}
%\end{figure}
Furthermore, we define  a ball area $\mathcal{B}(\bs{x}_O,R)$ located at $\bs{x}_O$ with radius $R=\bar{r}+r_o$ that includes the whole volume of the robotic team and the object (see Fig. \ref{fig:valid_transition}). Finally, the $\mathcal{M}$ static obstacles are defined as closed spheres described by $\pi_m=\mathcal{B}(\bs{p}_{\pi_m},r_{\pi_m}),~m\in\{1,\ldots, \mathcal{M}\}$, where $\bs{p}_{\pi_m}\in \mathbb{R}^3$ is the center and the $r_{\pi_m}$ the radius of the obstacle $\pi_m$.  

\subsection{Navigation Function}\label{NF:ref}
The calculation of the desired object trajectory within the workspace $\mathcal{W}$ relies on the Navigation Function concept originally proposed by Rimon and Koditschek in \cite{Koditschek1990412} as follows: \vspace{-0mm}
  \begingroup\makeatletter\def\f@size{8.9}\check@mathfonts
\def\maketag@@@#1{\hbox{\m@th\large\normalfont#1}}\begin{align}
\phi_O(\bs{x}_O;\bs{x}^d_O)=\frac{\gamma(\bs{x}_O-\bs{x}^d_O)}{[\gamma^k(\bs{x}_O-\bs{x}^d_O)+\beta(\bs{x}_O)]^{\frac{1}{k}}}\label{eq8}
\end{align}\endgroup
where   \begingroup\makeatletter\def\f@size{8.5}\check@mathfonts
\def\maketag@@@#1{\hbox{\m@th\large\normalfont#1}}$\phi_O:\xrightarrow{\mathcal{W}-\overset{\mathcal{M}}{\underset{m=1}{\cap}}\mathcal{B}(\bs{p}_{\pi_m},r_{\pi_m})}[0,1)$\endgroup denotes the potential that derives a safe motion vector field within the free space $\mathcal{W}-\overset{\mathcal{M}}{\underset{m=1}{\cap}}\mathcal{B}(\bs{p}_{\pi_m},r_{\pi_m})$. Moreover, $k>1$ is a design constant, $\gamma(\bs{x}_O-\bs{x}^d_O)\!>\!0$ with $\gamma(\bs{0})\!=\!0$ represents the attractive potential field to the goal position $\bs{x}^d_O\!$ and $\beta(\bs{x}_O)\!>\!0$ \cite{Koditschek1990412}: \vspace{-0mm}
  \begingroup\makeatletter\def\f@size{9.5}\check@mathfonts
\def\maketag@@@#1{\hbox{\m@th\large\normalfont#1}}\begin{equation*}
\lim\limits_{\bs{x}_O\rightarrow \begingroup\makeatletter\def\f@size{6}\check@mathfonts	
	\arraycolsep=0.4pt\def\arraystretch{0.6} \Bigg\{
	\begin{array}{ccc}
	\text{Boundary} \\
	\text{Obstacles}
	\end{array} 
	\endgroup}   \beta(\bs{x}_O)=0
\end{equation*}\endgroup \vspace{-4mm}

\noindent represents the repulsive potential field by the workspace boundary and the obstacle regions. It was proven in \cite{Koditschek1990412} that $\phi_O(\bs{x}_O,\bs{x}^d_O)$ has a global minimum at $\bs{x}^d_O$ and no other local minima for sufficiently large $k$. Thus, a feasible path that leads from any initial obstacle-free configuration to the desired configuration might be generated by following the negated gradient of $\phi_O(\bs{x}_O,\bs{x}^d_O)$. Consequently, the object's desired motion profile is designed as follows: \vspace{-0mm}
  \begingroup\makeatletter\def\f@size{9.5}\check@mathfonts
\def\maketag@@@#1{\hbox{\m@th\large\normalfont#1}}\begin{align}
{\bs{v}}^d_{O}(t)=-K_{NF}{\bs{J}'_O({\bs{\eta}}_{2,O})}\nabla_{\bs{x}_O}\phi_O(\bs{x}_O(t),\bs{x}^d_O)\label{eq9}
\end{align}\endgroup \vspace{-4mm}

\noindent where $K_{NF}>0$ is a positive gain. Now let us define a sequence of sampling time $\{t_j\}_{j\geq0}$ with a constant sampling time $h>0$ with $h<T_p$ for the system such that: \vspace{-1mm}
  \begingroup\makeatletter\def\f@size{9.5}\check@mathfonts
\def\maketag@@@#1{\hbox{\m@th\large\normalfont#1}}\begin{align}
t_{j+1}=t_j+h,~ \forall j\geq0\label{sampling_time}
\end{align}\endgroup \vspace{-5mm}

\noindent Therefore, given a current position and velocity of the object $\bs{x}_O(t_j)$, $\bs{v}_O(t_j)$ at the time $t_j$ each UVMS can propagates for time interval $s\in [t_j,t_j+T_P]$ where $T_P$ is the prediction horizon, a map of desired trajectory and velocity of the object based on \eqref{eq8}, \eqref{eq9} given as $\bs{x}^d_{O}(s)$ and $\bs{v}^d_{O}(s),~ s\in [t_j,t_j+T_P]$ which will be used in the subsequent analysis. 
%\begin{remark}
%	Notice that we assume that all UVMS $i\in\mathcal{N}$ are aware of both the desired configuration of the object as well as of the obstacles position in the workspace. Thus, propagating for time interval $t\in [t_i,t_i+T_P]$  based on \eqref{eq8} and \eqref{eq9} 
%\end{remark}

\subsection{Constraints} \vspace{0mm}
\textbf{State Constraints:}\\
We consider for each UVMS a set of constraints which are captured by the state constraint set $X_i$ of the system, given by:\vspace{-3mm}
  \begingroup\makeatletter\def\f@size{9.5}\check@mathfonts
\def\maketag@@@#1{\hbox{\m@th\large\normalfont#1}}\begin{align}
\bs{x}_i(t)\in X_i\subset \mathbb{R}^{2n_i}\label{state_const}
\end{align}\endgroup \vspace{-5.5mm}

\noindent which is formed by the following constraints:
  \begingroup\makeatletter\def\f@size{9.5}\check@mathfonts
\def\maketag@@@#1{\hbox{\m@th\large\normalfont#1}}\begin{subequations}
	\begin{gather}
	\theta_O(t)\in (-\frac{\pi}{2},\frac{\pi}{2})\label{eq5a}\\
	\bs{q}_i\in\mathbb{R}^{n_i} \backslash \Big({Q}_{s_i}(\bs{q}_i)\cup Q_{l_i}(\bs{q}_i)\Big),~i\in\mathcal{N}\label{eq5b}\\
	|\dot{{q}}_{k_i}|\leq \bar{\dot{{q}}}_i, \forall k\in \{1,\ldots,n_i\},i\!\in\!\mathcal{N} \label{eq5d}
	\end{gather}
\end{subequations}\endgroup
 where ${Q}_{s_i}(\bs{q}_i)$ is the set of singular position of the system \eqref{singular_robot} and $Q_{l_i}(\bs{q}_i)$ is the set of manipulator's joint limits given:\vspace{-1mm}
  \begingroup\makeatletter\def\f@size{9.5}\check@mathfonts
\def\maketag@@@#1{\hbox{\m@th\large\normalfont#1}}\begin{align}
\!\!Q_{l_i}\!(\bs{q}_i)\!=\!\{\bs{q}_i\!\in\!\mathbb{R}^{n_i}\!\!:|{q}_{k_i}|\!\leq\! \bar{{q}}_{k_i}\!\},\forall k\in \{1,\ldots,n\},i\!\in\!\mathcal{N}\!  \label{Joint_limits}
\end{align}\endgroup  \vspace{-4mm}

\noindent where $\bar{{q}}_{k_i}$ is the limit bound for the corresponding joint ${q}_{k_i},~k\in \{1,\ldots,n\},i\!\in\!\mathcal{N}$. Moreover, $\bar{\dot{{q}}}_i$ is the upper value for the joint velocity $\dot{{q}}_{k_i},~k\in \{1,\ldots,n\},i\!\in\!\mathcal{N}$. Therefore,  the set $X_i$ capture all the state constraints of the systems \eqref{nominal_system}, i.e., singularity avoidance as well as joint limits limitations. \vspace{1mm}
%\begin{remark}
%Notice that collision avoidance between the whole system (UVMS and the object) and obstacles within the workspace (see Fig \ref{fig:valid_transition}) are achieved based on the desired trajectory and velocity of the object calculated from  \eqref{eq8} and \eqref{eq9} as it is explained previously. 
%\end{remark} \vspace{3mm}

\textbf{Input Constraints:}\\
We consider the input constraints for each UVMS as: \vspace{-2mm}
  \begingroup\makeatletter\def\f@size{9.5}\check@mathfonts
\def\maketag@@@#1{\hbox{\m@th\large\normalfont#1}}\begin{gather*}
||\bs{\tau}_i||\leq\bar{\bs{\tau}}_i\Leftrightarrow||\bs{J}_i(\bs{q}_i)^\top\bs{u}_i ||\leq\bar{\bs{\tau}}_i
\end{gather*}\endgroup \vspace{-5mm}

\noindent where $\bar{\bs{\tau}}_i$ is a vector including corresponding limit bound for each actuated joint ${\tau}_{k_i}$, $k\in \{1,\ldots,\tau_{n_i}\},i\!\in\!\mathcal{N}$ where $\tau_{n}$ is the number of actuated joints. Therefore, we can define the control input set $T_i$: \vspace{-2.5mm}
  \begingroup\makeatletter\def\f@size{9.5}\check@mathfonts
\def\maketag@@@#1{\hbox{\m@th\large\normalfont#1}}\begin{align}
\bs{\tau}_i(t)\in T_i \subset \mathbb{R}^{\tau_{n_i}}\label{control_set}
\end{align}\endgroup \vspace{-6mm}

\noindent with: \vspace{-3mm}
  \begingroup\makeatletter\def\f@size{9.5}\check@mathfonts
\def\maketag@@@#1{\hbox{\m@th\large\normalfont#1}}\begin{gather*}
T_i=\{ \bs{\tau}_i\in \mathbb{R}^{\tau_{n_i}}:||\bs{J}_i(\bs{q}_i)^\top\bs{u}_i ||\leq\bar{\bs{\tau}}_i,~ \forall \bs{x}_i\in X_i \}
\end{gather*}\endgroup
\subsection{Control design}
At each sampling time, the UVMS $i \in \mathcal{N}$ solves an NMPC scheme subject to its corresponding dynamics \eqref{nominal_system} and a number of inequality constraints (i.e., \eqref{eq5a}-\eqref{eq5d} and \eqref{control_set}) in order to follow the computed desired trajectory $\bs{x}^d_{O}(s)$ and velocity $\bs{v}^d_{O}(s)$ for a time interval $s\in[t_j,t_j+T_P]$ based on \eqref{eq8}, \eqref{eq9} and \eqref{sampling_time}. In particular, in sampled data NMPC, a Finite Horizon Optimal Control Problem (FHOCP) is solved at discrete sampling time instants $t_j$ based on the current state measurements $\bs{x}_i(t_j),~i\in \mathcal{N}$. For UVMS $i,~i\in\mathcal{N}$, the open-loop input signal applied in between the sampling instants is given by the solution of the FHOCP: \vspace{-2mm}
\begingroup\makeatletter\def\f@size{9.5}\check@mathfonts
\def\maketag@@@#1{\hbox{\m@th\large\normalfont#1}}\begin{subequations}
	\begin{align}&\min_{\hat{\bs{\tau}}_i(\cdot)}J_i(\bs{x}(t_j),\hat{\bs{\tau}}_i(
	\cdot))=\label{12a}\\ 
	&\min_{\hat{\bs{\tau}}_i(\cdot)}\Bigg\{\int_{t_i}^{t_i+T_p}\Big[F_i\big(\hat{\bs{x}}_O(s),\hat{\bs{v}}_O(s),\hat{\bs{\tau}}_i(s)\big)\Big]ds\nonumber\\
	&\qquad\qquad+E_i\Big(\hat{\bs{x}}_O(t_j+T_P),\hat{\bs{v}}_O(t_j+T_P)\Big)\Bigg\}\nonumber\\
	&\text{subject to:}\nonumber\\
	&\hat{\dot{\bs{x}}}_i(s)=f_i(\hat{\bs{x}}_i(s),\hat{\bs{u}}_i(s)),\quad~\hat{\bs{x}}_i(t_j)=\bs{x}_i(t_j),\label{12bb}\\
	&\hat{\bs{\tau}}_i(s)=\bs{J}_i^\top(\hat{\bs{q}}_i)\hat{\bs{u}}_i+\bs{\tau}_{i0}(\bs{q}_i),~s\in[t_j,t_j+T_P]\\
	& \hat{\bs{x}}_O(s)=\mathcal{F}(\hat{\bs{q}}_i(s))-\left[
	\begin{array}{ccc}
	^I\bs{R}_O \bs{l}_i\\
	\boldsymbol{\alpha}_i\end{array} \right],~s\in[t_j,t_j+T_P],\label{12b}\\
	&\hat{\bs{v}}_O(s)=\bs{J}_{i_O}\bs{J}_i(\hat{\bs{q}}_i(s))\hat{\dot{\bs{q}}}_i(s),\qquad s\in[t_j,t_j+T_P],\label{12c}\\
	&\hat{\bs{x}}_i(s)\in X_i, \qquad \,\,\,\,s\in[t_j,t_j+T_P],\\
	&\hat{\bs{\tau}}_i(s) \in T_i,  \qquad s\in[t_j,t_j+T_P],\label{12f}\end{align}
\end{subequations}\endgroup \vspace{-5mm}

\noindent where $F$ and $E$ are the running and terminal cost function respectively which are both of quadratic form i.e., $F(\cdot)=\hat{\bs{x}}_O^\top\bs{Q}_x\hat{\bs{x}}_O+\hat{\bs{v}}_O^\top\bs{Q}_v\hat{\bs{v}}_O+{\bs{\tau}_i}^\top\bs{R}{\bs{\tau}_i}$ and $E(\cdot)=\hat{\bs{x}}_O^\top\bs{P}_x\hat{\bs{x}}_O$, respectively, with $\bs{P}_x$, $\bs{Q}_x$, $\bs{Q}_v$ and $\bs{R}$ being positive definite matrices to be appropriately tuned \cite{heshmati2018robust_icra}.  {\begin{comment}The disturbance term $\bs{w}_i$ can cause discrepancies between the predicted state given from the model \eqref{nominal_system} subject to a specific trajectory of inputs and the actual state, given from \eqref{actual_dynamics}  for the same trajectory of inputs.\end{comment}} 
In order to distinguish the predicted variables (i.e., internal to the controller) we use the double subscript notation $(\hat{\cdot})$ corresponding to the system \eqref{12bb}. This means that $\hat{\bs{x}}_i(s),~s\in[t_j,t_j+T_P]$ is the solution of  \eqref{nominal_system} based on the measurement of the state at time instance $t_j$ (i.e., $\bs{x}_i(t_j)$) while applying a trajectory of inputs (i.e., $\hat{\bs{u}}_i(s),~s\in[t_j,t_j+T_P]$).  
%The cost function $F_i(\cdot)$, as well as the terminal cost $E(\cdot)$, are both of quadratic form given as:\vspace{-2mm}
%
%{\small{\begin{gather*}
%F\!_i\!\big(\!\hat{\bs{x}}_O\!(s),\!\hat{\bs{v}}_O\!(s),\!\hat{\bs{\tau}}\!_f\!(s)\big)\!\!=\!\![\hat{\bs{x}}^\top_O\!(s)\!,\!\hat{\bs{v}}^\top_O\!(s)]\bs{Q}[\hat{\bs{x}}^\top_O\!(s)\!,\!\hat{\bs{v}}^\top_O\!(s)]\!^\top\!\!\!\!+\!\hat{\bs{\tau}}^\top_f\!\!(s)\bs{R}\hat{\bs{\tau}}\!_f\!(s)\\
%E_i\big(\hat{\bs{x}}_O,\!\hat{\bs{v}}_O\big)=[\hat{\bs{x}}^\top_O(s),\hat{\bs{v}}^\top_O(s)]\bs{P}[\hat{\bs{x}}^\top_O(s),\hat{\bs{v}}^\top_O(s)]^\top
%\end{gather*}}}\vspace{-2mm}
\noindent The solution of FHOCP \eqref{12a}-\eqref{12f} at time $t_j$ provides an optimal control input trajectory denoted by $\hat{\bs{\tau}}^*_i(s;\bs{x}(t_j)),~s\in[t_j,t_j+T_P]$. This control input is then applied to the system until the next sampling time $t_{j+1}$: i.e., $\bs{\tau}_i(s;\bs{x}(t_j))=\hat{\bs{\tau}}^*_i(s;\bs{x}(t_j)),~s\in[t_j,t_j+h]
$. 
At time $t_{j+1}=t_j+h$ a new finite horizon optimal control problem is solved in the same manner, leading to a receding horizon approach. Notice that the control input $\bs{\tau}_i(\cdot)$ is of feedback form, since it is recalculated at each sampling instant based on the then-current state.  
\section{Simulation study}
Real-time simulation have been performed to validate the proposed approach. The simulation environment is designed based on UwSim dynamic simulator \cite{UWSIM} running on the Robot Operating System (ROS) \cite{ROS}. We consider a scenario involving 3D motion with two UVMSs  with the same structure, transporting a bar in a constrained workspace with static obstacles (see Fig.\ref{fig:UVMS}). The UVMS model is an AUV equipped with a small $4$ DoF manipulator attached at the bow of the vehicle (see Fig.\ref{fig:UVMS}). The dynamic parameters of the vehicle have been identified via a proper identification scheme \cite{karras2018unsupervised}, while the manipulator's parameters as well as object's parameters have been extrapolated by the CAD data. The complete state vector of the vehicle (3D position, orientation, velocity) is available via the sensor fusion and state estimation module given in our previous results \cite{karras2018unsupervised}. The Constrained NMPC employed in this work is implemented using the NLopt Optimization library \cite{nlopt}.
\begin{figure}[th]
	%	\graphicspath{ {Initial/} }
	\centering
	\subfloat[]{%
		\includegraphics[width=0.47\linewidth]{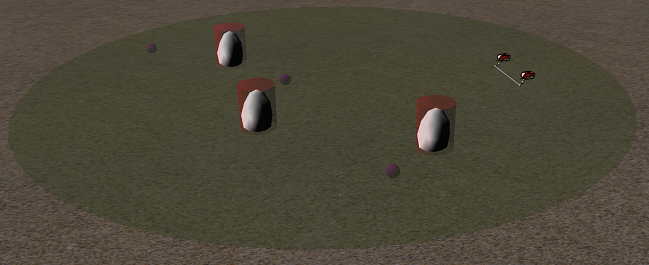}}
	\label{1a}\hfill
	\subfloat[]{%
		\includegraphics[width=0.47\linewidth]{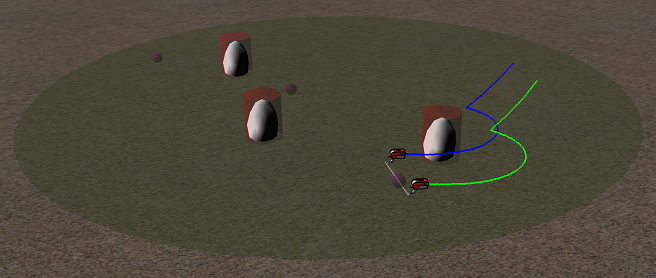}}
	\label{1b}\\
	\subfloat[]{%
		\includegraphics[width=0.47\linewidth]{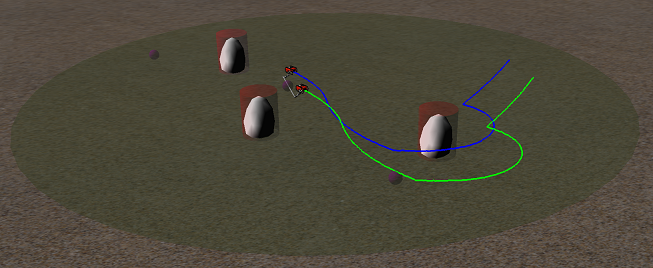}}
	\label{1c}\hfill
	\subfloat[]{%
		\includegraphics[width=0.47\linewidth]{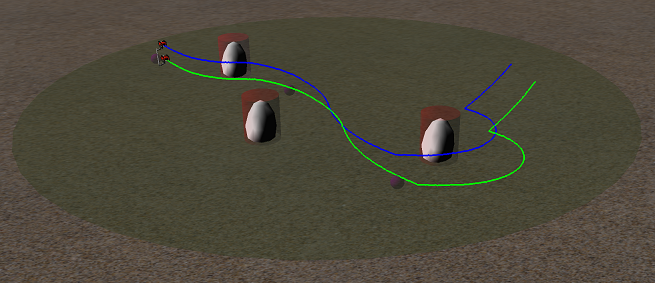}}
	\label{1d} 
%	\hfill
%	\subfloat[]{%
%		\includegraphics[width=0.46\linewidth]{Screenshot12sep_14.png}}
%	\label{1e} 
%	\hfill
%	\subfloat[]{%
%		\includegraphics[width=0.46\linewidth]{Screenshot12sep_18.png}}
%	\label{1f} 
%	\hfill
%	\subfloat[]{%
%		\includegraphics[width=0.45\linewidth]{Screenshot12sep_14.png}}
%	\label{1g} 
%	\hfill
%	\subfloat[]{%
%		\includegraphics[width=0.45\linewidth]{Screenshot12sep_18.png}}
%	\label{1h} 
\vspace{0mm}	\caption{ The evolution of the proposed methodology in 4 consecutive time
		instants.}\vspace{-2mm}
	\label{figure6} 
\end{figure}
\begin{figure}
	\centering
		\includegraphics[width=3.4in]{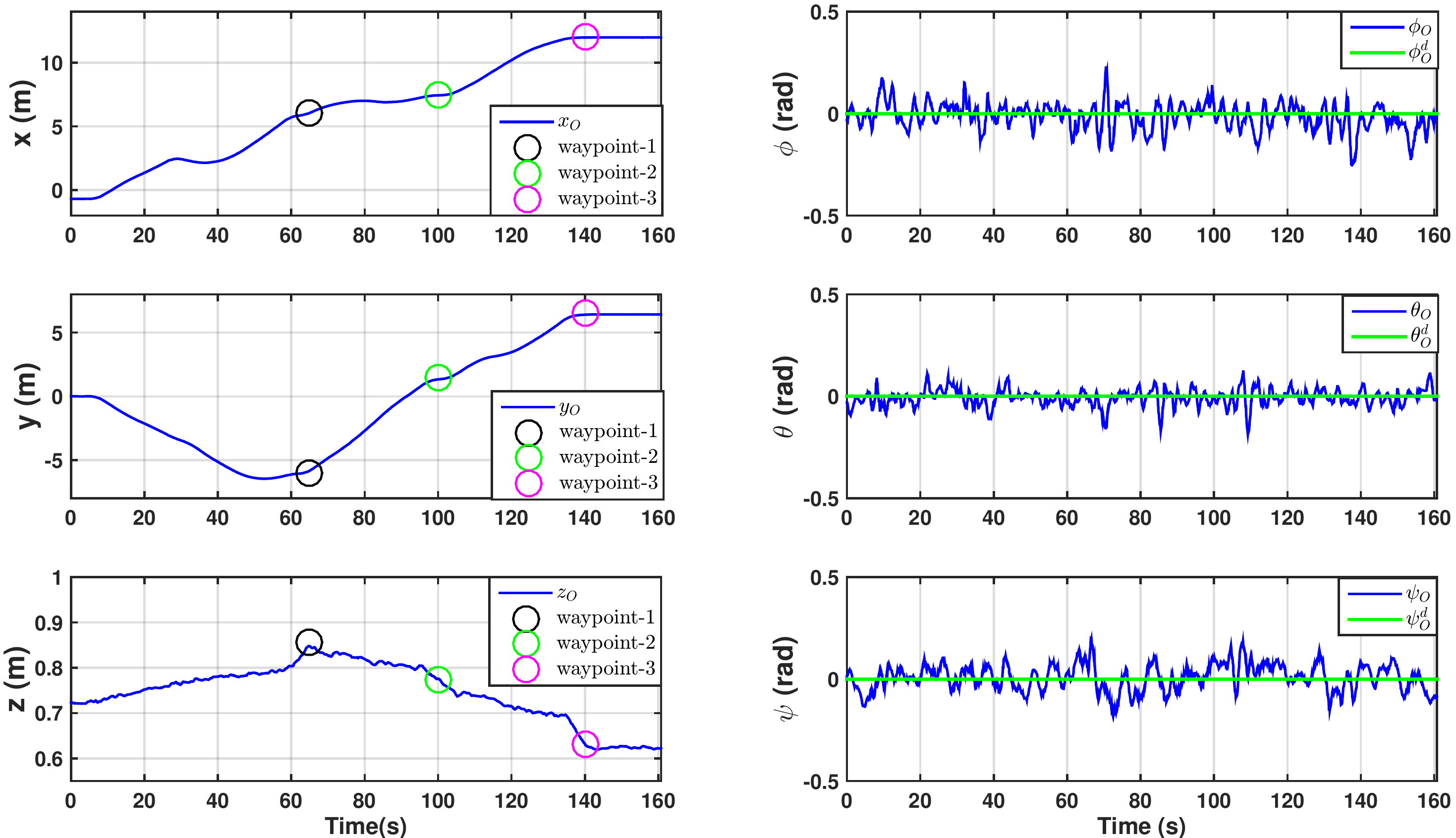}\vspace{0mm}
\vspace{0mm}	\caption{Object coordinates during the control operation \vspace{0mm} }\label{object_coor_traj}\vspace{-2mm}
\end{figure}
\begin{figure}
	\centering
	\vspace{0mm}	\includegraphics[width=3.0in]{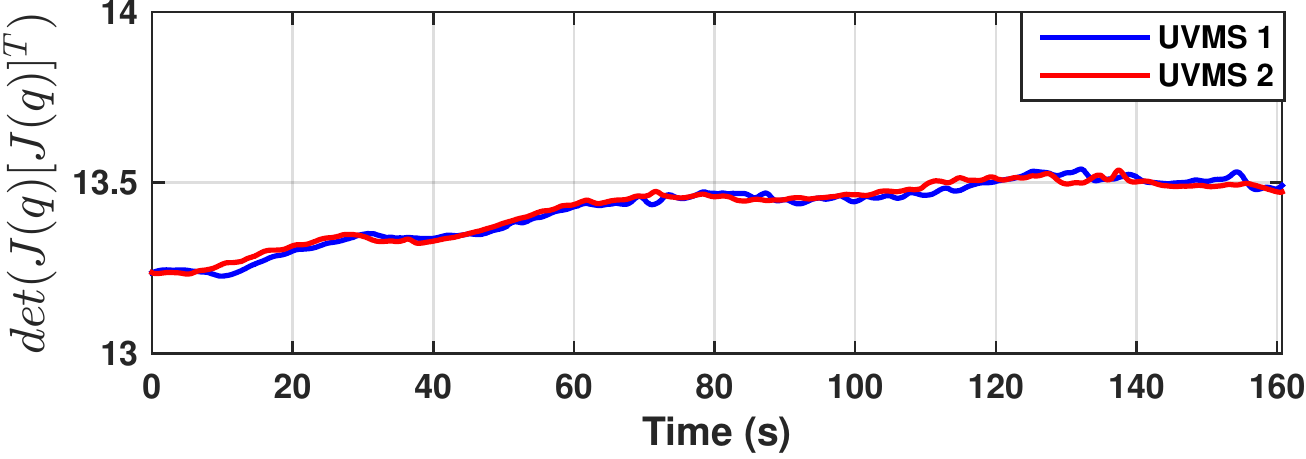}\vspace{0mm}
	\vspace{0mm}\caption{ $det(\bs{J}(\bs{q})[\bs{J}(\bs{q})]^\top)$ during the control operation \vspace{0mm} }\label{det_jac}\vspace{-2mm}
\end{figure}
\begin{figure}
	\centering
	\vspace{0mm}	\includegraphics[width=3.2in]{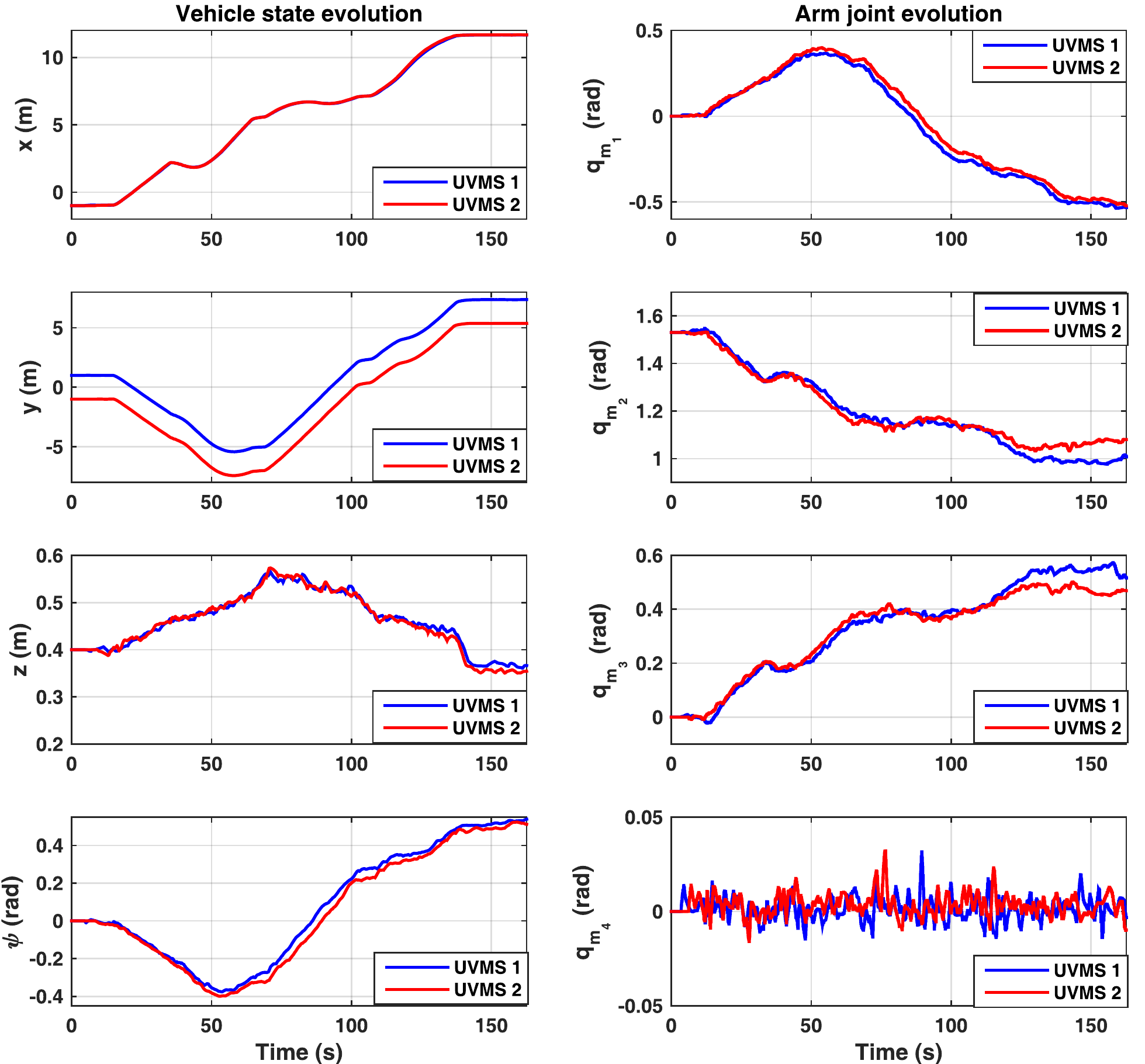}\vspace{0mm}
	\vspace{-0mm}\caption{The evolution of the system states at joint level  \vspace{0mm} }\label{stetes_robot}\vspace{-0mm}
\end{figure}
\begin{figure}
	\centering
	\vspace{0mm}	\includegraphics[width=3.2in]{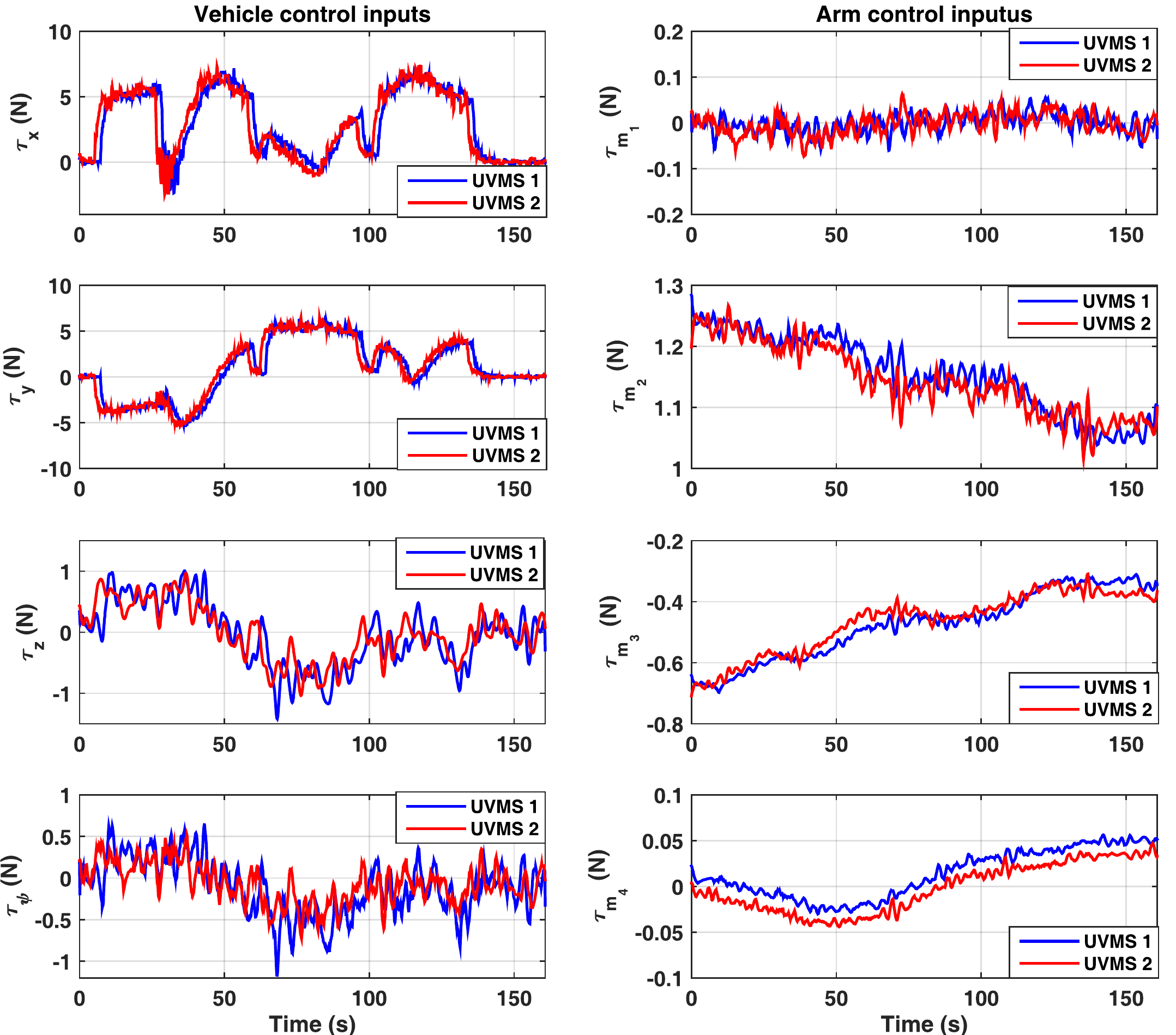}\vspace{0mm}
	\vspace{-0mm}\caption{The control input signals during the control operation \vspace{0mm} }\label{control_robot}\vspace{-2mm}
\end{figure}
\begin{figure}
	\centering
	\vspace{0mm}	\includegraphics[width=3.3in]{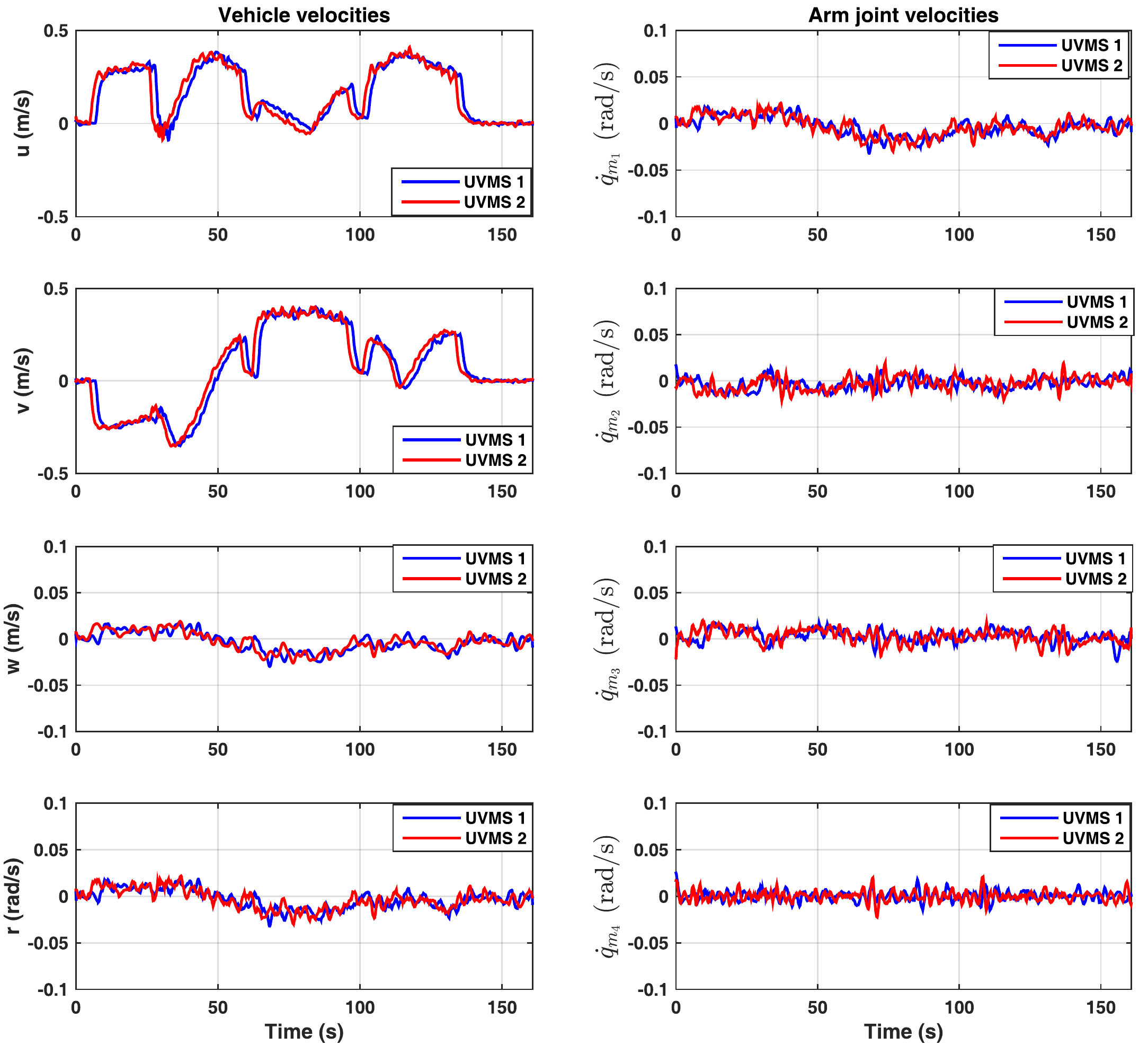}\vspace{0mm}
\vspace{-0mm}	\caption{The evolution of the system velocities at joint level \vspace{0mm} }\label{velocity}\vspace{-2mm}
\end{figure}
\subsection{Simulation results}
In the following simulation, the objective for the team of UVMSs is to follow a set of predefined way points, while simultaneously avoid obstacles within the workspace. The position of the obstacles w.r.t the inertial frame $\mathcal{I}$ in $x-y$ plane is given by: $\bs{x}_{{obs}_1}=[4,~-4.5]$, $\bs{x}_{{obs}_2}=[9,~-1.5]$ and $\bs{x}_{{obs}_3}=[9,~5]$ respectively.  These obstacles are cylinders with radius $r_{\pi_i}=0.6m,~ i=\{1,2,3\}$ and are modeled together with the workspace boundaries according to the spherical world representations as consecutive spheres.  The radius of the sphere $\mathcal{B}(\bs{p}_i,\bar{r}),~i\in\{1,2\}$ which covers all the UVMS volume (for all possible configurations) is defined as $\bar{r}=1 m$. In this way, the Navigation function \eqref{eq8}-\eqref{eq9} was designed with gain $K_{NF}=0.5$. Regarding to constraints \eqref{eq5d}, we consider that the vehicle's velocity  most not exceed $0.5m/s$}for translation and $0.1rad/s$ for rotational.  In the same vein, the manipulator joint velocities must be retained between $(-0.1,0.1)rad/s$. Moreover, the manipulator joint positions  \eqref{Joint_limits} must be retained between  $(-2,2)rad$. Furthermore, input saturations \eqref{control_set} for the vehicle and manipulator are considered as: $\bar{\tau}_v=10 N$  and $\bar{\tau}_m=2 N$,  respectively. The sampling time \eqref{sampling_time}  and the prediction horizon are  $h=0.12 sec$ and $T_p=5\times h= 0.6 sec$ respectively. The matrices $\bs{P}_x$, $\bs{Q}_x$, $\bs{Q}_v$ and $\bs{R}$ as well as he load sharing coefficients $c_1$ and $c_2$ for both UVMSs are equal and set to: $\bs{P}_x=\bs{Q}_x=0.8\cdot\bs{I}_{6\times6}$, $\bs{R}=0.3\cdot\bs{I}_{8\times8}$, $\bs{Q}_v=0.4\cdot\bs{I}_{6\times6}$, and $c_1=c_2=0.5$. The initial position of the object is $\bs{x}_{{O}}=[-0.7,~0,~0.72,~0.04,~-0.07,~0]$. We set $3$ waypoints as $\bs{x}_{{O}_1}^d=[6,~-6,~0.85,~0,~0,~0]$, $\bs{x}_{{O}_2}^d=[7.5,~1.5,~0.78,~0,~0,~0]$ and $\bs{x}_{{O}_3}^d=[12,~6.5,~0.65,~0,~0,~0]$ which make the mission more challenging considering the obstacles' positions within the workspace (See Fig.\ref{figure6} and Fig.\ref{fig:UVMS}). The results are presented in Fig.\ref{figure6}-Fig.\ref{det_jac}. The trajectory of the system within the workspace as well as object coordinates evolution are depicted in Fig. \ref{figure6} and Fig.\ref{object_coor_traj} respectively. It can be seen that the UVMSs have successfully transported cooperatively the object and have followed the set of predefined way points while safely avoids the obstacles. The evolution of $det(\bs{J}(\bs{q})[\bs{J}(\bs{q})]^\top)$ (see \eqref{singular_robot} and \eqref{eq5b}) during the operation is given in Fig.\ref{det_jac}. It can be easily seen that value remained positive during the cooperative manipulation task. Moreover, the evolution of the system velocity and its states at joints level as well as the corresponding control inputs are indicated in  Fig.\ref{velocity}, Fig. \ref{stetes_robot} and Fig.\ref{control_robot} respectively. As it was expected from the theoretical findings, these values were retained in the corresponding feasible regions defined by the corresponding upper bounds and consequently all of the system constraints were satisfied.
\section{Summary and Future Work}
In this paper we presented a novel distributed object transportation control scheme for a team UVMSs in a constrained workspace with static obstacles. Various limitation and constraints such as: obstacles, joint limits, control input saturation as well as kinematic and representation singularities have been considered during the control design. The proposed control strategy relieves the team of robots from intense inter-robot communication during the execution of the collaborative tasks. This, avoids any restrictions imposed by the acoustic communication bandwidth (e.g., the number of participating UVMSs). Moreover, the control scheme adopts load sharing among the UVMSs according to their specific payload capabilities. Future research efforts will be devoted towards experimental validation of the proposed methodology with two small UVMSs inside a small test tank. In the same spirit, considering non-holonomic constraints on the UVMS model is a promising direction that would increase the applicability of the proposed control scheme.

%
%\begin{figure}[t!]
%	\centering
%	\setlength{\fboxsep}{0pt}%
%	\setlength{\fboxrule}{2pt}%
%	\fbox{\includegraphics[width=0.45\textwidth]{UVMS_coop.jpg}}
%	\caption{Custom-made UVMSs under cooperative transportation.}
%	\label{ch:9:fig:uvms_frames}
%\end{figure}

%In this paper, we presented a distributed adaptive object transportation scheme for Underwater Vehicle Manipulator Systems under implicit communication, avoiding thus completely tedious explicit data transmission. In the proposed scheme, only the leading UVMS  aware of the desired configuration of the object and  the obstacles' position in the workspace, and aims at navigating the overall formation towards the goal configuration while avoiding collisions with the static obstacles. On the contrary, the followers adopt a prescribed performance estimation technique in order to estimate the object's desired trajectory. Each following UVMS employs the proposed estimator based on its own local measurements, thus the estimated desired trajectory of the object for each following UVMS is relative to its own inertial frame. In this way, the whole team mutually agrees on a desired trajectory of the commonly grasped object. Moreover, contrary to the existing work in the related literature, the proposed scheme imposes no restrictions on the underwater communication bandwidth. 
%
% Future research efforts will be devoted towards extending the proposed methodology for multiple UVMSs with underactuated vehicle dynamics and incorporating various load sharing policies based on the payload capabilities of the UVMSs.

\bibliographystyle{ieeetr}
\bibliography{mybibfilealina}
\end{document}